\documentclass[letterpaper]{article} 
\usepackage{aaai2026}  
\usepackage{times}  
\usepackage{helvet}  
\usepackage{courier}  
\usepackage[hyphens]{url}  
\usepackage{graphicx} 
\usepackage{natbib}  
\usepackage{caption} 
\usepackage{algorithm}
\usepackage{algorithmic}

\usepackage{newfloat}
\usepackage{listings}
\DeclareCaptionStyle{ruled}{labelfont=normalfont,labelsep=colon,strut=off} 
\floatstyle{ruled}
\newfloat{listing}{tb}{lst}{}
\floatname{listing}{Listing}
\usepackage{todonotes}

\newcommand{\mypara}[1]{\smallskip\noindent\textbf{#1.} \xspace}

\usepackage[strict]{changepage}
\usepackage{framed}
\definecolor{customshade}{rgb}{0.941, 0.937, 0.996}
\definecolor{darkblue}{rgb}{0.14,0.22,0.62}
\newenvironment{kkboxline}{%
  \MakeFramed{\advance\hsize-\width\FrameRestore}%
  \noindent\hspace{-4.55pt}
  \begin{adjustwidth}{}{7pt}%
}
{%
  \end{adjustwidth}\endMakeFramed%
}

\definecolor{StripeGray}{gray}{0.95}    
\definecolor{lightgray}{gray}{0.92}

\definecolor{starfill}{HTML}{F7CFCE}
\definecolor{staroutline}{HTML}{FF7D78}

\newcommand{\customstar}{%
    \begin{tikzpicture}[baseline=-0.6ex]
        \node[
            star,
            star points=5,
            star point ratio=2.2,
            fill=starfill,
            draw=staroutline,
            line width=0.3pt,
            inner sep=0pt,
            minimum size=7pt
        ] {};
    \end{tikzpicture}%
}
\usepackage{enumitem}
\usepackage{xspace}
\usepackage{amsfonts}
\usepackage{amsmath}

\usepackage{amssymb}
\usepackage{tikz}
\usepackage{tcolorbox}
\usepackage{enumitem}
\usepackage{xcolor}
\usepackage{booktabs}

\usepackage{booktabs}
\usepackage{xcolor,colortbl}
\usepackage{siunitx}
\usepackage{tabularray}

\usepackage{booktabs}
\usepackage{siunitx}
\usepackage{multirow}
\usepackage{makecell}

\usepackage{scalefnt,sectsty}

\usepackage{colortbl}

\usepackage{caption}
\usepackage{subcaption}

\usepackage{xcolor}
\usepackage{tikz}
\usetikzlibrary{shapes.geometric}

\title{Evaluating the Dynamics of Membership Privacy in Deep Learning}
\author{
    Yuetian Chen\textsuperscript{\rm 1},
    Zhiqi Wang\textsuperscript{\rm 2},
    Nathalie Baracaldo\textsuperscript{\rm 3},
    Swanand Ravindra Kadhe\textsuperscript{\rm 3},
    Lei Yu\textsuperscript{\rm 4}
}
\affiliations{
    \textsuperscript{\rm 1}Purdue University\\
    \textsuperscript{\rm 2}Pennsylvania State University\\
    \textsuperscript{\rm 3}IBM Research\\
    \textsuperscript{\rm 4}Rensselaer Polytechnic Institute
}

\begin{document}

\maketitle
\begin{abstract}
  Membership inference attacks (MIAs) pose a critical threat to the privacy of training data in deep learning. Despite significant progress in attack methodologies, our understanding of when and how models encode membership information during training remains limited. This paper presents a dynamic analytical framework for dissecting and quantifying privacy leakage dynamics at the individual sample level. By tracking per-sample vulnerabilities on an FPR-TPR plane throughout training, our framework systematically measures how factors such as dataset complexity, model architecture, and optimizer choice influence the rate and severity at which samples become vulnerable. Crucially, we discover a robust correlation between a sample's intrinsic learning difficulty, and find that the privacy risk of samples highly vulnerable in the final trained model is largely determined early during training.
Our results thus provide a deeper understanding of how privacy risks dynamically emerge during training, laying the groundwork for proactive, privacy-aware model training strategies.

\end{abstract}
\section{Introduction}
\label{sec:intro}

Membership Inference Attacks (MIAs)~\cite{shokri2017membership} represent a significant threat to privacy in deep learning, enabling adversaries to infer whether a specific data point was used in model training. Beyond their immediate implications for individual privacy~\cite{long2018understanding,yeom2018privacy,lira,choquette2021label,watson2022importance}, MIAs also play a critical role as tools for auditing differentially private systems~\cite{236254,9519424}, verifying machine unlearning~\cite{bourtoule2020machineunlearning,10.1145/3460120.3484756}, and quantifying the privacy risks of model memorization~\cite{mireshghallah2022quantify}. Despite considerable advances in developing increasingly sophisticated attacks, our fundamental understanding of how, when, and why membership information becomes encoded in models during the training process remains surprisingly limited.

\begin{figure}[h]
    \centering
    \includegraphics[width=\linewidth]{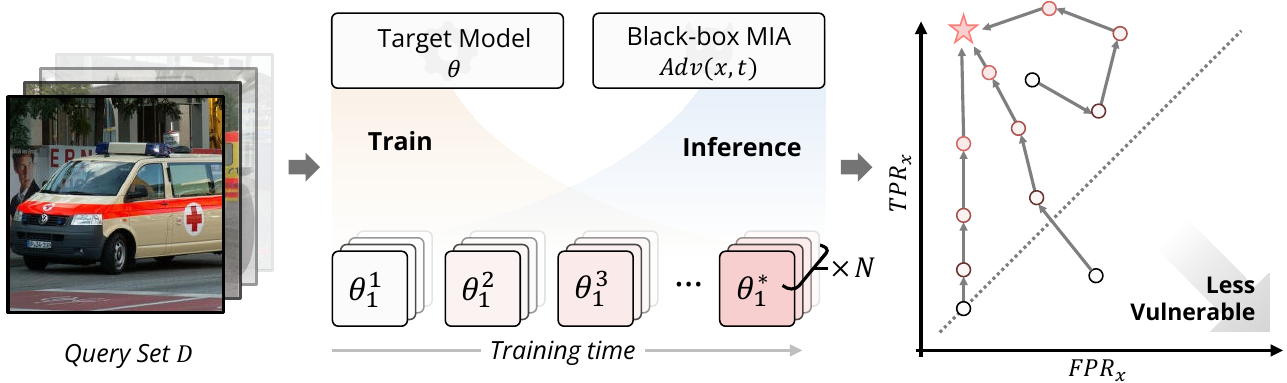}
    \caption{\small \textbf{An illustration of sample-level vulnerability trajectories.} Each dot on a path represents a sample’s vulnerability state, characterized by its membership attack metric (FPR, TPR), estimated from $N$ model variations at a given epoch. The distinct paths illustrate how membership encoding evolves uniquely for each sample---nuances obscured by traditional post-hoc analyses that only assess the final state (\customstar).}
    \label{fig:overview}
    \vspace{-10pt}
\end{figure}

Current research primarily relies on post-hoc assessments of trained models~\cite{ye2022enhanced,277402,chen2024hideplainsightcleanlabel}, evaluating privacy vulnerabilities only after training completion. Although insightful, this static approach provides merely a snapshot of vulnerability at a single point in time (i.e, the final training epoch), ignoring the underlying dynamic processes through which privacy risks emerge and evolve. Consequently, it provides little insight into the dynamics of sample-level privacy leakage throughout training and offers limited guidance for proactively preventing privacy risks during the model training phase.


To bridge this critical gap, we present a novel dynamic analytical framework designed to systematically dissect and quantify membership privacy leakage at the granularity of individual training samples. Central to our approach is the concept of tracking per-sample vulnerabilities over time on an FPR-TPR vulnerability plane, as illustrated in Figure~\ref{fig:overview}, allowing us to visualize and measure the evolving privacy risks throughout the entire training process. This dynamic, sample-centric perspective provides critical insights into how various training-time factors, including dataset complexity, model architecture, and optimizer choice, influence the rate and extent to which membership information is encoded into models. While previous works~\cite{yeom2018privacy,nasr2019comprehensive,long2018understanding,lira} have explored the relationship between membership inference vulnerabilities and these factors, their analyses have predominantly been static, post-training evaluations focusing on qualitative assessments of final vulnerabilities. In contrast, our approach quantitatively measures the dynamic impact of these factors on membership privacy leakage throughout training, significantly advancing beyond existing static methodologies.



Specifically, our contributions are threefold:
\begin{enumerate}[leftmargin=*, topsep=0pt, partopsep=0pt, itemsep=2pt]
    \item We propose a new framework and a suite of metrics, including membership encoding speed, center of mass displacement, and transition probabilities, to characterize the dynamic nature of privacy loss for individual samples and entire populations. The source code of our framework will be made publicly available.
    \item Through extensive experiments, we present—for the first time—a systematic and quantitative characterization of the dynamic impact of dataset complexity, model architecture, and optimizer choice, based on per-sample membership encoding trajectories. Our approach reveals how and when privacy vulnerabilities emerge and evolve during training—a significant advancement beyond existing static, qualitative analyses.
    \item We reveal a strong correlation between intrinsic sample hardness—quantified through metrics such as influence functions, learning iterations, and epistemic uncertainty—and membership vulnerability. Crucially, our analysis demonstrates that vulnerability of these “hard-to-learn” samples is predominantly determined early during training, indicating a critical window for proactive intervention. Moreover, we show that optimization methods explicitly promoting generalization, notably Sharpness-Aware Minimization (SAM)~\cite{foret2021sharpnessawareminimizationefficientlyimproving}, substantially mitigate the encoding of membership information throughout training.
\end{enumerate}

Ultimately, our work shifts the perspective from reactive, post-hoc privacy auditing toward proactively characterizing how and when privacy risks emerge during training. By providing analytical tools and empirical insights into the dynamics of privacy leakage, our framework lays critical groundwork for developing targeted interventions and more effective privacy-aware training strategies.

\section{Related Work}

\mypara{Membership Inference Attacks}
Membership Inference Attacks aim to determine if a specific data record was used to train a model, a vulnerability extensively documented across various architectures~\citep{shokri2017membership, truex2019demystifying, hu2022membership, rigaki2023survey, leino2020stolen}. Foundational approaches demonstrated that models behave differently for members versus non-members, often using shadow models to learn these distinguishing patterns~\citep{salem2018mlleaks, hayes2017logan, choquettechoo2021labelonly}. The state-of-the-art has since evolved to include more robust, reference-based methods that calibrate a target's outputs against a baseline~\citep{jayaraman2019evaluating, mireshghallah2022quantify, song2021systematic, sablayrolles2019white}. Among these, the Likelihood Ratio Attack (LiRA) is particularly effective at providing reliable, instance-specific scores~\citep{lira}.

\mypara{Learning Dynamics as a Privacy Lens}
Existing research has demonstrated that the learning process itself is a primary source of information leakage, primarily driven by model overfitting~\citep{yeom2018privacy, long2018understanding, jayaraman2020revisiting, zanella2020analyzing}. A rich body of work connects a model's vulnerability to its learning dynamics—the temporal evolution of its state and parameters. Specific indicators of memorization and heightened risk have been identified through signals like loss and prediction confidence trajectories~\citep{li2021membership, song2019privacy}, patterns of example ``forgetting"~\citep{toneva2018empirical, feldman2020does}, and the information content revealed by gradient evolution~\citep{carlini2021extracting, nasr2019comprehensive, hitaj2017deep}. Gradient information is a particularly potent leakage channel, directly exploited by severe attacks that can reconstruct training data~\citep{zhu2019deep, geiping2020inverting, zhao2020idlg}. Furthermore, this granular perspective can uncover disparate privacy risks, where specific data subgroups become significantly more vulnerable due to algorithmic bias or reliance on spurious correlations~\citep{bagdasaryan2019differential, chang2021privacy, suriyakumar2021chasing}. While these studies link individual dynamic events to privacy risk, they often analyze them as static properties. For instance, a sample's entire loss history will be averaged into its single final value, or its complex pattern of being forgotten and re-learned is simplified to a total forgetting count. This post-hoc approach discards crucial temporal information about \emph{how} and \emph{when} vulnerability emerges—the central question our framework addresses by analyzing the full vulnerability trajectory over time.

\section{Threat Model}
\label{sec:threat_model}
This section formally defines the MIA threat model underlying our analytical framework. Our goal is not to introduce a new attack; rather, we leverage a robust, established attack paradigm as a measurement tool to systematically analyze how and when privacy vulnerabilities emerge during the training process.

\noindent\textbf{Information Exposure Model.}
We define a \textbf{black-box setting}, aligned with most previous works~\cite{shokri2017membership,yeom2018privacy,jayaraman2020revisiting,ye2022enhanced,lira}. The adversary can query the model $\mathcal{M}_\theta$ with any input $x'$ and receive its detailed output vector $f_{\mathcal{M}_\theta}(x') \in \mathbb{R}^C$, containing class logits or posterior probabilities. For our dynamic analysis, this query access extends to a sequence of model checkpoints $\{\mathcal{M}_{\theta_t}\}_{t=1}^T$, where $\theta_t$ represents the parameters after $t$ training epochs. 
Conversely, the adversary is completely walled off from the model's internal state and training procedure. This concealed information includes all model parameters $\{\theta_t\}$, any gradients computed during training $\nabla_{\theta_t} \mathcal{L}$, the full training set $D_{train}$ itself, and specific training hyperparameters.

\noindent\textbf{The MIA Game.}
Following previous work~\cite{yeom2018privacy, lira}, we formalize MIAs as a game between an adversary and a challenger. Let the training dataset $D_{train}$ and a disjoint holdout dataset $D_{out}$ be drawn i.i.d. from an underlying data distribution $\mathcal{D}$. A target model $\mathcal{M}_{\theta}$ is trained on $D_{train}$. A challenger flips a random bit $b \in \{0, 1\}$ and selects a sample $x$ from $D_{train}$ if $b=1$, or from $D_{out}$ if $b=0$. The adversary $\mathcal{A}$, given black-box access to $\mathcal{M}_{\theta}$ and the target sample $x$, must output a guess $b'$. The adversary wins the game if $b' = b$. The success of the attack is quantified by the adversary's advantage:

\begin{align}
\small
    \text{Adv}(\mathcal{A}, \mathcal{M}_{\theta}) ={} & \Pr_{x \sim D_{\text{train}}}[\mathcal{A}(x, \mathcal{M}_{\theta}) = 1]  \nonumber \\
    &  - \Pr_{x \sim D_{\text{out}}}[\mathcal{A}(x, \mathcal{M}_{\theta}) = 1]
\end{align}

This advantage is equivalent to the difference between the attack's True Positive Rate (TPR) and False Positive Rate (FPR), which characterizes how well an adversary can distinguish between members and non-members. A positive advantage implies that membership information can be exploited by the adversary.

\noindent\textbf{Dynamic Analysis Perspective.}
Our threat model follows standard assumptions used by established reference-based MIAs such as LiRA~\cite{lira}, but introduces a dynamic dimension. Specifically, our framework leverages the exposed information to compute the sample-level membership advantage score, $\text{Adv}(x, t)$, at each training checkpoint $t$. Rather than conducting static, post-training audits, our approach focuses on analyzing the resulting vulnerability trajectory ($\text{Adv}(x, 1)$, \dots, $\text{Adv}(x, T)$). This dynamic perspective enables a detailed exploration of how membership information is progressively encoded into models over the entire training process.
\section{A Dynamic Framework for Membership Privacy Analysis}
\label{sec:framework}
The conventional approach of evaluating privacy risks using a single, post-hoc measurement on a fully trained model provides an incomplete view, obscuring the underlying mechanisms and temporal dynamics of information leakage. To address this limitation, we introduce a framework that explicitly models membership privacy as a dynamic, evolving process. Our framework is built upon a formal, sample-centric definition of vulnerability, represented geometrically in a state-space, and supported by a suite of metrics derived from stochastic process theory. Together, these components enable precise characterization of how privacy risks evolve throughout training.

\subsection{A Formalization for Per-Sample Vulnerability}

Let $\mathcal{Z} = \mathcal{X} \times \mathcal{Y}$ be the data space over which a probability distribution $\mathcal{D}$ is defined. A training algorithm $\mathcal{T}$ is a (potentially stochastic) function that maps a training set $D \subset \mathcal{Z}^n$ to a model parameterization $\theta \in \Theta$, where $\Theta$ is the parameter space. Our analysis focuses on the vulnerability of a specific target sample $z \in \mathcal{Z}$.

To formalize the privacy risk, we consider two hypotheses concerning the generation of a model $\mathcal{M}_\theta$: the \textit{in}-hypothesis, $H_1$, where $z$ is included in the training set, and the \textit{out}-hypothesis, $H_0$, where it is not. This induces two distributions over the parameter space $\Theta$. Given a random dataset $S \sim \mathcal{D}^{n-1}$, we define the \textit{in}-distribution as $\mathcal{P}_{\mathrm{in}} = \mathcal{A}(S \cup \{z\})$ and the \textit{out}-distribution as $\mathcal{P}_{\mathrm{out}} = \mathcal{A}(S)$.
A membership inference attack $\mathcal{F}: \Theta \times \mathcal{Z} \to \{0, 1\}$ is a statistical test designed to distinguish between these two distributions based on the model's behavior on $z$. The efficacy of this test for a specific sample $z$ is captured by its sample-level membership advantage.

\begin{tcolorbox}[
    colback=gray!3!white,
    colframe=gray!75!black,
    boxrule=0.5pt,
    arc=0pt,
    left=8pt,
    right=8pt,
    top=6pt,
    bottom=6pt
]
\textbf{Definition 1 (Sample-Level Membership Advantage).}
For an attack $\mathcal{F}$ and sample $z$, the advantage is the statistical difference in the attack's success probability under the two hypotheses:
\begin{equation}
\label{eq:adv_def}
\mathrm{Adv}^{\mathcal{F}}(z) = \mathbb{E}_{\theta \sim \mathcal{P}_{\mathrm{in}}}[\mathcal{F}(\theta, z)] -  \mathbb{E}_{\theta \sim \mathcal{P}_{\mathrm{out}}}[\mathcal{F}(\theta, z)]
\end{equation}
\end{tcolorbox}

This metric is estimated empirically via a Monte Carlo simulation. We generate a population of $N$ models, $\{\mathcal{M}_i\}_{i=1}^N$, where for each model, the inclusion of $z$ is governed by a Bernoulli random variable $b_i \sim \mathrm{Bernoulli}(0.5)$. This allows us to compute the empirical True Positive Rate ($\mathrm{TPR}_{z}$) and False Positive Rate ($\mathrm{FPR}_{z}$) over the model population. The empirical advantage $\widehat{\mathrm{Adv}}^{\mathcal{F}}(z)  = \mathrm{TPR}_{z} - \mathrm{FPR}_{z}$, denoted by $\alpha_{z}$, serves as our fundamental instantaneous measure of the privacy risk for the sample $z$.



\subsection{Characterizing Membership Encoding Dynamics}
\label{subsec:characterizing_membership_encoding_dynamics}
To characterize the dynamics of membership encoding, we introduce a framework that analyzes privacy leakage at two distinct scales: the individual sample and the entire population. This dual-level approach allows us to first model the specific vulnerability path of each data point and then aggregate these individual behaviors to understand the collective, system-level dynamics of how membership information becomes embedded in a model during training.

\subsubsection{Individual-level Sample Dynamics} 
Our analysis begins at the most granular level: the individual data sample. To quantify how a single sample's privacy risk evolves, we first map its vulnerability to a geometric state space, the ``vulnerability plane." By tracking the sample's position on this plane across training epochs, we can visualize its unique ``vulnerability trajectory" and measure key dynamic properties, such as the speed at which it becomes susceptible to inference.

\mypara{The Vulnerability Plane as a Geometric State Space}
To characterize the dynamics of membership privacy, we represent a sample's vulnerability state as a point $v(z) = (\mathrm{FPR}_{z}, \mathrm{TPR}_{z})$ within an FPR-TPR plane, termed \emph{vulnerability plane}.  This geometric space is defined as the unit square $\mathcal{V} = [0, 1] \times [0, 1]$, where the $x$ and $y$-axes correspond to FPR and TPR, respectively. Within this plane, the advantage $\alpha$ can be viewed as a scalar field evaluated at each point $v(z)$, specifically given by $\alpha(z) = \mathrm{TPR}_{z} -\mathrm{FPR}_{z}$.

Using this vulnerability plane, the training process induces a \emph{vulnerability trajectory} for each sample $z$, which is a sequence of states $\tau(z) = (v_1(z), v_2(z), \ldots, v_T(z))$ in $\mathcal{V}$, where each state $v_i(z)$ is estimated empirically from a population of $N$ models trained for $i$ epochs, and $T$ is the total number of training epochs.
Based on it, our analysis introduces a hierarchy of metrics to formally characterize the dynamics of this process, viewing the evolution of the sample population as a stochastic system.

\mypara{Membership Encoding Speed} To measure the membership encoding speed for a sample, we compute its velocity over its vulnerability trajectory, defined as $\vec{s}(z, t) = v_{t+1}(z) - v_t(z) \in \mathbb{R}^2$. This vector represents the discrete derivative of a sample's vulnerability state, capturing the precise epoch-wise changes at its $(\mathrm{FPR}_{z}, \mathrm{TPR}_{z})$ position.

\subsubsection{Population-level Sample Dynamics}
We introduce a suite of metrics to model and analyze the collective evolution of sample vulnerabilities in a general setting. These metrics capture both homogeneous and heterogeneous properties of membership encoding behavior across samples.

\mypara{Transition Matrix} 
We discretize the vulnerability plane $\mathcal{V} = [0, 1]^2$ into a finite set of disjoint states. Specifically, in this paper, we partition the vulnerability plane into nine disjoint states by applying a 3$\times$3 grid with diving both TPR and FPR axes to intervals $I_1 = [0, 1/3)$, $I_2 = [1/3, 2/3)$, and $I_3 = [2/3, 1]$ Each state $S_{ij} = (I_i^{\text{TPR}}, I_j^{\text{FPR}}$) represents the cell $(i,j)$ on the grid. We model the population dynamics as a time-inhomogeneous Markov process. The transition probabilities between states at each time step $t$ are captured by a transition matrix $A(t)$, whose entries $a_{ij,kl}(t)$ represent the fraction of samples transitioning between $S_{ij}$ and $S_{kl}$ from epoch $t$ to $t+1$. The resulting sequence $\{A(t)\}_{t=1}^{T-1}$ quantitatively captures the temporal dynamics of privacy risk evolution, including critical transitions such as from a non-vulnerable state (e.g., $S_{11}$) to a high-risk state (e.g., $S_{31}$) that has a high TPR and low FPR.

Furthermore, we consider the positions of all $M$ samples on the vulnerability plane, i.e., $\{v_t(z_j)\}_{j=1}^M$, where $v_t(z_j) = (\mathrm{FPR}_{z_j}, \mathrm{TPR}_{z_j})$ represents the vulnerability coordinates of sample $z_j$ at epoch $t$. To systematically characterize the distribution of these vulnerability positions, we define a suite of aggregate metrics that capture their motion dynamics, information entropy, and clustering structure.

\mypara{Motion Metrics} We quantify the overall drift of the population's average vulnerability using the Center of Mass (CoM) Displacement. The CoM at epoch $t$ is the mean of all sample coordinates, $\mathrm{CoM}_t = \frac{1}{M} \sum_{j=1}^M v_t(z_j)$. The total displacement, computed as the cumulative path length $\sum_{t=1}^{T-1} ||\mathrm{CoM}_t - \mathrm{CoM}_{t-1}||_2$, measures the magnitude of the collective shift in privacy risk. We complement this metric with the Average Membership Encoding Speed, defined as the mean magnitude of velocity vectors $||\vec{s}(z, t)||_2$ across all samples and epochs, capturing the average rate at which individual sample vulnerabilities change over time.
Furthermore, we identify the dominant axis of the population's movement by computing a Directional Angle ($\theta$). This angle is derived from the principal eigenvector of the covariance matrix of the CoM's velocity field, indicating whether the collective drift is systematically oriented, for example, towards higher $\mathrm{TPR}$ or lower $\mathrm{FPR}$.
    
\mypara{Information Metrics} Following state discretization for transition matrix, we use Spatial Entropy to measure the heterogeneity of the vulnerability distribution across different states, $H(t) = -\sum_{s \in \mathcal{S}} p_s(t) \log p_s(t)$, where $p_s(t)$ is the fraction of samples in state $s$ at epoch $t$. We measure the total change in Spatial Entropy ($\Delta H = H(T) - H(0)$) to quantify the net diversification of risk profiles over training.

\mypara{Clustering Metrics} To analyze how samples with similar vulnerabilities form groups, we apply a density-based clustering algorithm at each epoch using the samples’ vulnerability coordinates. We quantify the typical degree of fragmentation in the vulnerability landscape through the Average Number of Clusters ($\bar{k}$), calculated across all epochs. Additionally, we introduce the Change in Clusters ($\Delta k$), defined as the difference between the number of clusters at the final and initial epochs. This metric indicates whether groups of similarly vulnerable samples tend to merge or subdivide as training progresses, highlighting the evolution of distinct vulnerability profiles over the training process.

\section{Evaluation}
In this section, we use our dynamic framework and metrics to empirically examine and quantify how dataset complexity, model architecture, and optimizer choice affect the temporal dynamics of membership encoding.

\subsection{Experimental Setup}
\noindent\textbf{Datasets.} We use four standard image classification datasets, selected to represent a spectrum of increasing complexity: MNIST, Fashion-MNIST (F-MNIST), CIFAR-10, and CINIC-10~\cite{mnist, fmnist, cifar10, darlow2018cinic10}. This allows us to study how feature and class complexity impact privacy dynamics.

\noindent\textbf{Models and Training.} Our experiments utilize several Convolutional Neural Network (CNN)~\cite{lecun1998gradient} architectures of varying depths and a deeper Wide Residual Network (WRN)~\cite{zagoruyko2016wide}, specifically \texttt{wrn28-2}. 
We employ three optimizers in the experiment. For general experiments, such as comparing datasets and architectures, we use the popular adaptive method AdamW~\cite{loshchilov2017decoupled}. In the experiment specifically analyzing the impact of optimizers, we include Sharpness-Aware Minimization (SAM)~\cite{foret2021sharpnessawareminimizationefficientlyimproving} in particular because it explicitly seeks wider and flatter minima to enhance generalization, which may inherently reduce model memorization and, consequently, mitigate privacy leakage. AdamW shows the impact similar to the standard SGD with momentum~\cite{ruder2016overview}, and thus we focus primarily on comparing SAM with standard SGD to quantify how optimizer choice influences the dynamics of membership privacy.
For each experimental configuration, we save model checkpoints at 10-epoch intervals throughout a 400-epoch training process to enable a fine-grained temporal analysis.

\noindent\textbf{MIA Method.} To instantiate our framework, we require a robust method to estimate the per-sample advantage, $\alpha_z$. We employ the Likelihood Ratio Attack (LiRA)~\cite{lira}, known for its strong performance, particularly in the low false-positive rate regime. For each estimation, we use a population of 20 shadow models. Following established methodology, we adopt a global variance estimate computed across these shadow models, a strategy shown to achieve robust attack performance even with as few as 16 shadow models\cite{lira}
While LiRA is used for the primary analysis due to its precision, we confirm that our core findings on privacy dynamics remain consistent when using other classic attack methodologies, as detailed in the Appendix.

\subsection{Impact of Training-Time Factors on Membership Encoding Dynamics}
We examine the evolution of per-sample and population-level vulnerability metrics to analyze how dataset complexity, model capacity, and optimizer choice influence privacy leakage dynamics. Our results demonstrate that these factors profoundly and predictably affect the speed, severity, and heterogeneity of privacy leakage, underscoring that the dynamics of the learning process itself are as critical to privacy outcomes as the final trained model.

\begin{figure}
    \centering
    \includegraphics[width=\columnwidth]{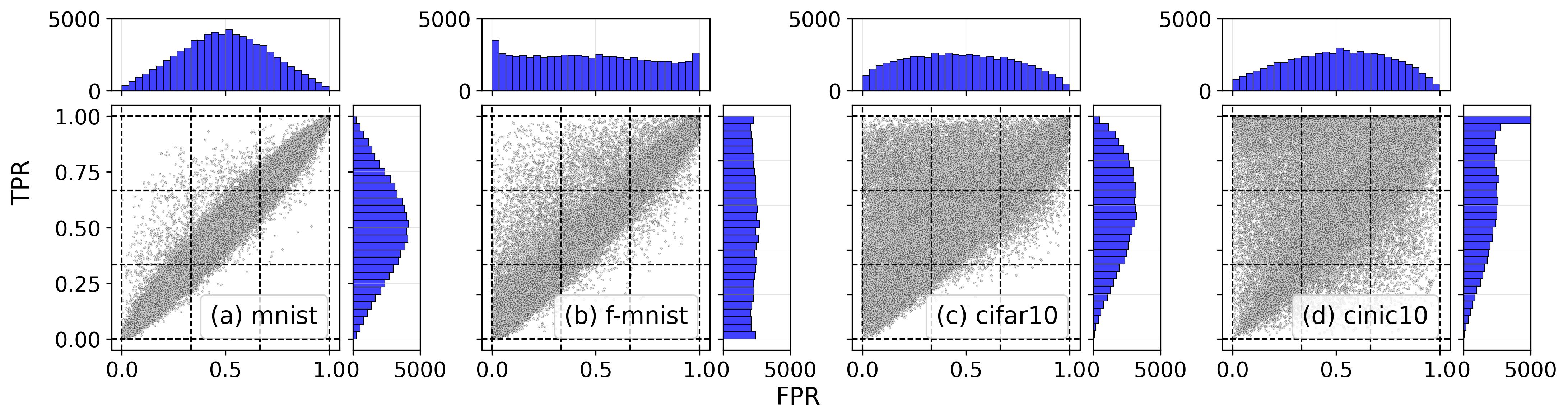}
    \caption{\small
Post-training FPR-TPR planes for different datasets: (a) MNIST, (b) Fashion MNIST, (c) CIFAR-10, and (d) CINIC-10. The histograms along two axes represent the density distributions of TPR and FPR values.}
    \label{fig:datasets_analyze}
\end{figure}

\subsubsection{Dataset Complexity}
\begin{kkboxline}
\textbf{Finding 1:}
\textit{
Dataset complexity is a primary driver of membership encoding dynamics; more complex datasets induce a faster, more extensive, and more heterogeneous encoding of vulnerable samples.
}
\end{kkboxline}

\noindent
\textbf{Description.}
We train a \texttt{wrn28-2} model on each of the four datasets—MNIST, Fashion-MNIST, CIFAR-10, and CINIC-10—which represent a spectrum of increasing complexity. We then apply our dynamic analysis framework to compare the evolution of their vulnerability trajectories and collective dynamics over 400 epochs of training.

\noindent
\textbf{Results.}
The intrinsic complexity of a dataset fundamentally shapes the model’s learning behavior, directly influencing its vulnerability to privacy leakage. 
For simpler datasets like MNIST, models can achieve low error by learning smooth, generalizable decision boundaries, resulting in limited privacy leakage. This effect is clearly visible in the empirical distribution of samples on the vulnerability plane (Figure~\ref{fig:datasets_analyze}). Specifically, samples that fall along the non-vulnerable diagonal (where FPR = TPR) reflect random-guess membership confidence, indicating minimal privacy risk. Consequently, models trained on simpler datasets typically exhibit sample distributions tightly concentrated near this diagonal. However, as dataset complexity increases, models must adapt by forming more intricate decision boundaries to handle higher intra-class variance and reduced inter-class separability. This adaptation often leads to the memorization of specific samples, evident from the progressive shift of the sample distribution toward the top-left, high-vulnerability region observed in more complex datasets such as CIFAR-10 and CINIC-10.




\begin{table}[t]
\centering
\small
\setlength{\tabcolsep}{6pt}
\rowcolors{1}{white}{gray!5}
\caption{Dynamic metrics across datasets (subscripts indicate standard deviation; values with $^{\dagger}$ scaled by $10^{-3}$).}
\label{tab:evaluation_metrics_data}
\begin{tabular}{@{}lcccc@{}}
\toprule
\textbf{Metric} & \textbf{MNIST} & \textbf{F-MNIST} & \textbf{CIFAR10} & \textbf{CINIC10} \\
\midrule
\multicolumn{5}{@{}l}{\textit{Motion}} \\
Speed ($\bar{v}$)$^{\dagger}$ & 0.38\textsubscript{0.42} & 1.28\textsubscript{0.77} & 2.95\textsubscript{1.30} & 3.50\textsubscript{1.30} \\
$\alpha{>}0$ Speed$^{\dagger}$ & 0.68\textsubscript{0.86} & 0.90\textsubscript{0.66} & 2.38\textsubscript{1.07} & 5.00\textsubscript{1.20} \\
CoM Disp. & 12.8 & 48.7 & 114.1 & 135.0 \\
\midrule
\multicolumn{5}{@{}l}{\textit{Information}} \\
Entropy ($H$) & 4.42\textsubscript{0.08} & 4.35\textsubscript{0.52} & 4.72\textsubscript{0.80} & 4.90\textsubscript{0.60} \\
$\Delta H$ & $-$0.08 & 1.47 & 3.47 & 4.20 \\
Dir. Angle & $-$0.77 & 1.05 & 0.96 & 1.00 \\
\midrule
\multicolumn{5}{@{}l}{\textit{Clustering}} \\
Avg. ($\bar{k}$) & 3132\textsubscript{39} & 2304\textsubscript{405} & 1741\textsubscript{445} & 1500\textsubscript{400} \\
$\Delta k$ & 6 & 1547 & 1282 & 1300 \\
\bottomrule
\end{tabular}
\end{table}
\begin{figure}
    \centering
    \includegraphics[width=1\columnwidth]{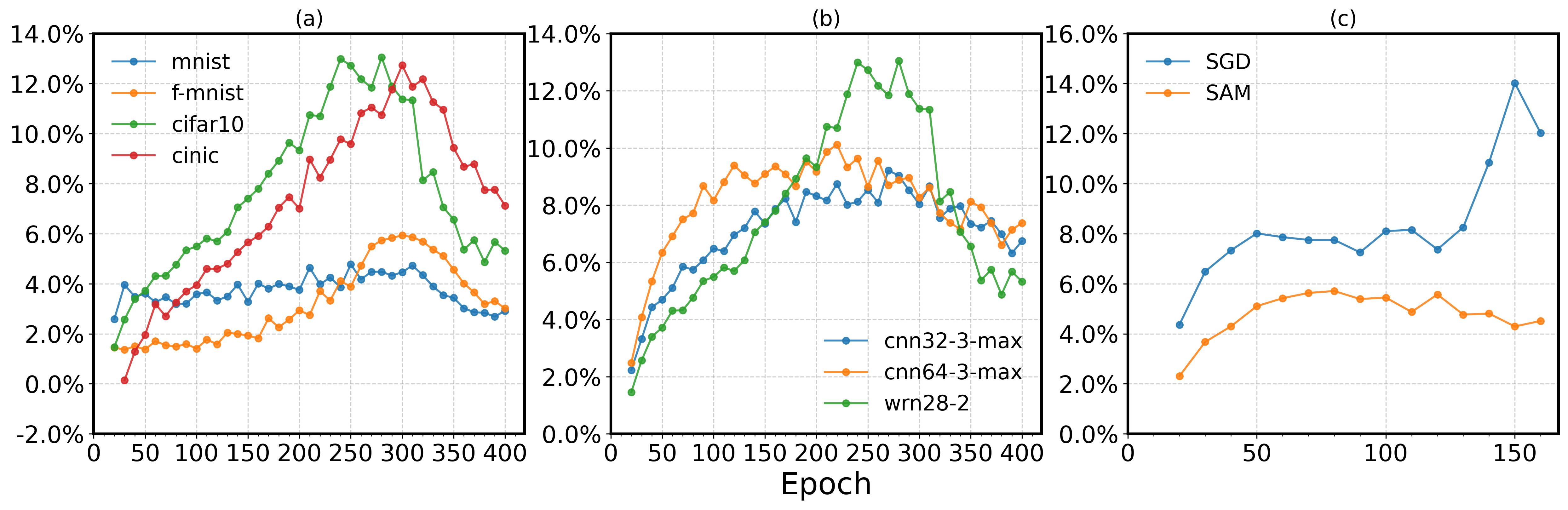}
    \caption{Transition probabilities from robust region(low-FPR, low-TPR) to highly vulnerable region across epochs for (a) different datasets, (b) model architectures, and (c) optimizers}
    \label{fig:trans_mat}
\end{figure}


Our collective dynamics metrics, summarized in Table~\ref{tab:evaluation_metrics_data}, rigorously quantify this phenomenon, highlighting a clear trend: increasing dataset complexity substantially amplifies both the magnitude and rate of privacy leakage. Specifically, the CoM Displacement increases more than tenfold, rising from 12.78 (MNIST) to 135.0 (CINIC-10), and the average Membership Encoding Speed nearly triples. This demonstrates that models trained on more complex datasets experience larger and faster parameter shifts, thereby embedding significantly greater membership information into the model during training.

Crucially, this leakage is not uniform. The substantial increase in Spatial Entropy change ($\Delta H$) for CIFAR-10 (3.47) and CINIC-10 (4.20) indicates that the population of samples bifurcates into distinct vulnerability groups during training. Initially, samples might reside in a relatively uniform, low-vulnerability state (low entropy). However, as training progresses, the sample population spreads across the vulnerability plane: some samples remain within the robust, easily generalizable region, while many others shift towards a highly vulnerable, memorized region. This divergence results in a “privacy-underprivileged” subset of samples, highlighting significant disparities that would otherwise remain hidden in aggregate privacy evaluations. Additionally, the Membership Encoding Speed for these vulnerable samples is notably higher, significantly shortening the available time window for effective privacy interventions.

Furthermore, we analyze the transition probability between two critical states: the \emph{Robust Region}, defined by low-FPR and low-TPR state $\boldsymbol{S_{11}}$, and the \emph{Highly Vulnerable Region}, defined by low-FPR and high-TPR state $\boldsymbol{S_{31}}$. Figure~\ref{fig:trans_mat} presents the transition probability $a_{11,31}(t) = P(v_{t+1} \in S_{31} \mid v_t \in S_{11})$ from the transition matrix, empirically capturing the fraction of samples that shift to a highly vulnerable state at epoch t. As illustrated in Figure~\ref{fig:trans_mat}(a), datasets with higher complexity demonstrate accelerated vulnerability dynamics. Specifically, a pronounced "tipping point" emerges, where the probability of transitioning to a high-risk state sharply rises, peaking at approximately 13\% for CIFAR-10.

\subsubsection{Model Architecture}
\begin{table}[t]
    \centering
    \footnotesize
    \setlength{\tabcolsep}{3pt}
    \caption{Metrics on different models (subscripts indicate standard deviation; values with $^{\dagger}$ scaled by $10^{-3}$).}
    \label{tab:evaluation_metrics_arch}
    \resizebox{\linewidth}{!}{%
    \begin{tabular}{@{}l>{\columncolor{gray!15}}lccc@{}}
        \toprule
        \textbf{Group} & \textbf{Metric} & \textbf{cnn32-3-max} & \textbf{cnn64-3-max} & \textbf{wrn28-2} \\
        \midrule
        \multirow{3}{*}{\textit{Motion}} & Speed ($\bar{v}$)$^{\dagger}$ & 1.10\textsubscript{0.68} & 2.85\textsubscript{1.20} & 2.95\textsubscript{1.30} \\
        & $\alpha{>}0$ Speed$^{\dagger}$ & 1.50\textsubscript{0.80} & 2.20\textsubscript{1.02} & 2.38\textsubscript{1.07} \\
        & CoM Disp. & 35.12 & 108.65 & 114.07 \\
        \midrule
        \multirow{3}{*}{\textit{Information}} & Entropy ($H$) & 4.40\textsubscript{0.20} & 4.69\textsubscript{0.75} & 4.72\textsubscript{0.80} \\
        & $\Delta H$ & 0.30 & 3.41 & 3.47 \\
        & Dir. Angle ($\theta$) & $-$0.50 & 0.93 & 0.96 \\
        \midrule
        \multirow{2}{*}{\textit{Clustering}} & Avg. ($\bar{k}$) & 2156\textsubscript{180.2} & 1710\textsubscript{420.3} & 1741\textsubscript{445.2} \\
        & $\Delta k$ & 820 & 1267 & 1282 \\
        \bottomrule
    \end{tabular}%
    }
    \vspace{-10pt}
\end{table}
\begin{kkboxline}
\textbf{Finding 2:}
\textit{
Deeper and more complex model architectures act as catalysts for membership encoding, accelerating the rate at which samples become vulnerable and creating more severe final privacy risks.
}
\end{kkboxline}

\noindent
\textbf{Description.}
We analyze three architectures of increasing complexity on the CIFAR-10 dataset: a 3-layer CNN with 32 filters (\texttt{cnn32-3-max}), a 3-layer CNN with 64 filters (\texttt{cnn64-3-max}), and the deeper \texttt{wrn28-2} model. We compare their collective dynamics metrics over the training period to isolate the impact of model capacity on the membership encoding process.

\noindent
\textbf{Results.}
It is well-known that models with larger capacity are generally prone to higher levels of memorization, thereby increasing their privacy risk. However, our dynamic framework captures a more nuanced insight: High-capacity models like \texttt{wrn28-2} aggressively memorize individual samples rather than being forced to find simpler, more generalizable solutions. The metrics presented in Table~\ref{tab:evaluation_metrics_arch} reveal a clear trend. The Center of Mass (CoM) Displacement for \texttt{wrn28-2} (114.07) is over three times larger than for the shallowest CNN (35.12). This signifies that the deeper model's parameters undergo far more substantial changes specifically to fit the training data, resulting in a much larger "privacy debt" accrued over training. This is not just a final-state phenomenon; the average Membership Encoding Speed ($\bar{v}$) is also highest for \texttt{wrn28-2}, indicating that this privacy debt is accrued more rapidly.



The most insightful metric is the change in Spatial Entropy ($\Delta H$). For the shallow \texttt{cnn32-3-max}, $\Delta H$ is minimal (0.30), suggesting its learning process is relatively uniform across samples. In stark contrast, `wrn28-2` has a $\Delta H$ of 3.47. This large increase in entropy demonstrates that the high-capacity model learns to create a highly heterogeneous risk landscape: it can perfectly generalize a subset of ``easy" samples (keeping them secure) while intensely memorizing a different subset of ``hard" samples, rendering them extremely vulnerable. Such a pronounced separation into a privacy-underprivileged subset is a distinctive characteristic of high-capacity models.

Figure~\ref{fig:trans_mat}(b) presents the transition probabilities for different models. Notably, the peak probability of samples transitioning from a robust to a vulnerable state is highest for the deeper \texttt{wrn28-2} model (13.9\%), compared to the shallower CNN models (10.3\% and 9.7\%). This indicates that deeper models are more inclined to undergo decisive, substantial changes in how they treat specific samples during critical training phases, consequently pushing these samples over the vulnerability threshold. 

\subsubsection{Optimizer Choice}
\begin{kkboxline}
\textbf{Finding 3:}
\textit{
The choice of optimizer critically influences the population's vulnerability trajectory, with Sharpness-Aware Minimization (SAM) actively suppressing the encoding of membership information.
}
\end{kkboxline}

\noindent
\textbf{Description.}
We compare the training dynamics of a \texttt{wrn28-2} model on CIFAR-10 using two different optimizers: standard Stochastic Gradient Descent (SGD) with momentum, and SAM. All other hyperparameters are held constant for a fair comparison over 160 epochs.

\begin{figure}[t!]
    \centering
    \begin{subfigure}[t]{0.32\columnwidth}
        \centering
        \includegraphics[width=\columnwidth]{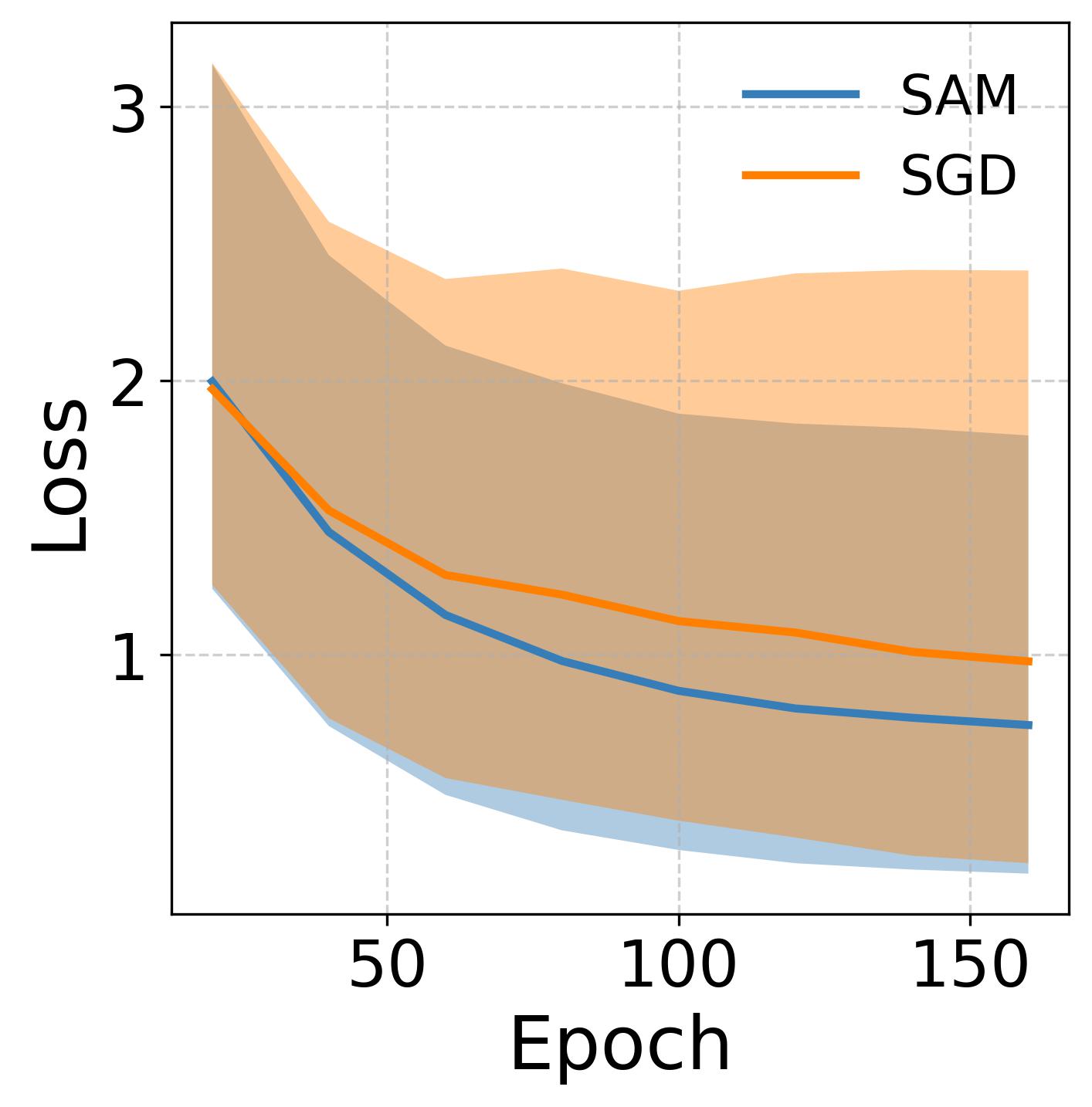}
        \caption{Loss progression during training with SGD and SAM.} 
        \label{fig:optim_loss}
    \end{subfigure}
    \hfill
    \begin{subfigure}[t]{0.65\columnwidth}
        \centering
        \includegraphics[width=\columnwidth]{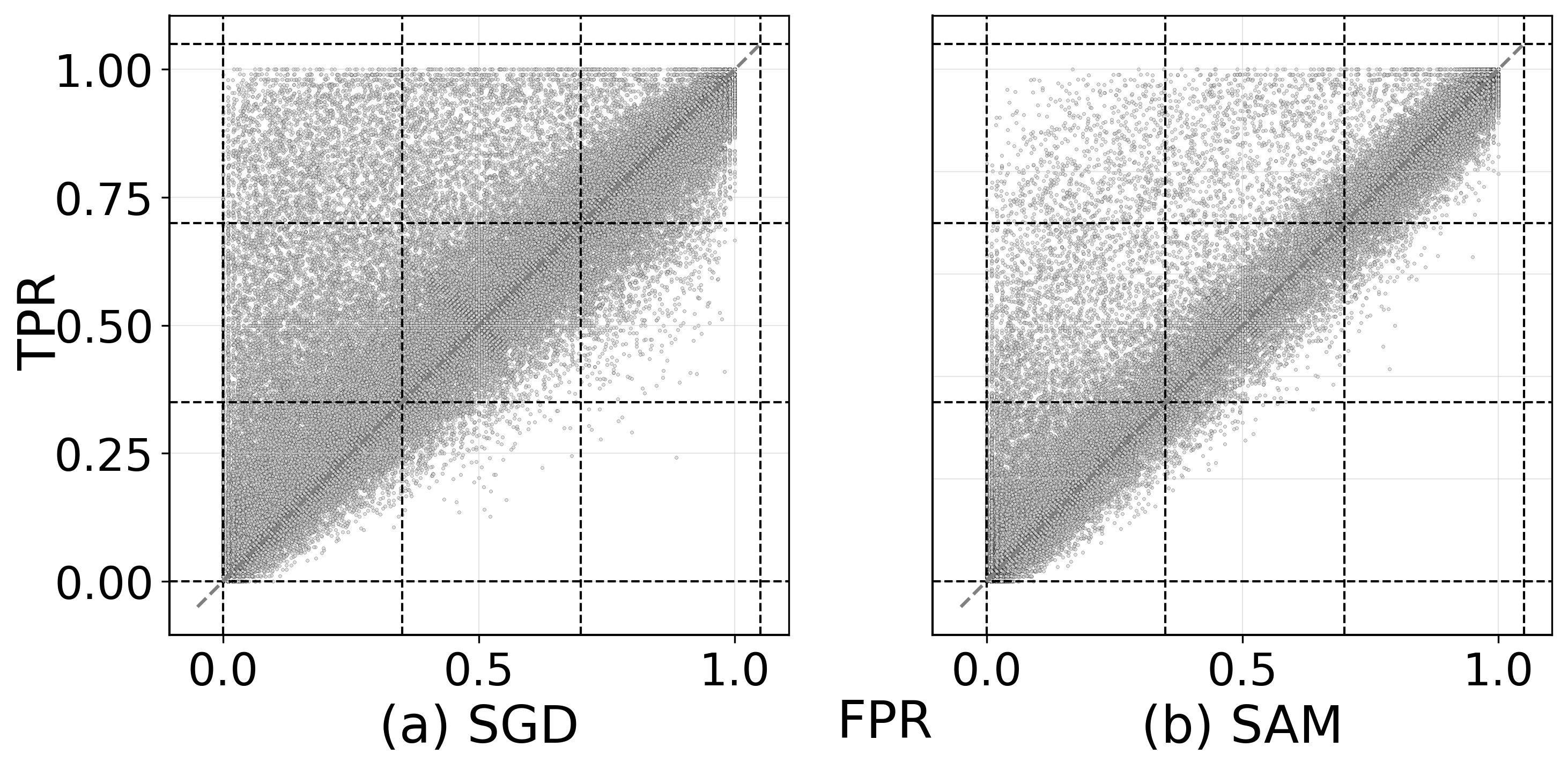}
        \caption{Post-training FPR-TPR plane under different optimizers: (a) SGD and (b) SAM.}
        \label{fig:optim_detect}
    \end{subfigure}%

    \caption{Comparison of training dynamics and detection performance under different optimizers.}
    \label{fig:datasets_analyze_score}
\end{figure}



\noindent
\textbf{Results.}
The choice of optimizer fundamentally influences the trajectory through the loss landscape, significantly impacting privacy. As visualized on the vulnerability plane in Figure~\ref{fig:optim_detect}, models trained with SAM exhibit markedly lower vulnerability, with their final distribution $\mu_T$ tightly clustered near the non-vulnerable diagonal. In contrast, SGD results in a dispersed distribution featuring many highly vulnerable samples, indicative of substantial memorization. This difference arises because SAM explicitly seeks wide, flat minima that encourage better generalization. Figure~\ref{fig:optim_loss} confirms this by illustrating SAM’s more stable and consistently lower loss profile, reducing the necessity for models to memorize individual samples to achieve low training error.

The dynamic analysis highlights an even more pronounced difference between optimizers. Specifically, as illustrated in Figure~\ref{fig:trans_mat}(c), the transition probability of samples moving into a highly vulnerable state peaks at 2.7\% for models trained with SGD, whereas for SAM-trained models, this peak is significantly lower at just 0.9\%. These results demonstrate that SAM not only achieves lower final vulnerability but actively mitigates the dynamic encoding of vulnerability throughout training by steering the optimization trajectory toward more generalizable regions of the parameter space.

\vspace{-10pt}
\subsection{Sample Hardness in Membership Privacy}
\vspace{-10pt}

\begin{kkboxline}
\textbf{Finding 4:}
\textit{
A sample's susceptibility to membership inference is strongly correlated with its learning difficulty. Metrics that quantify cumulative model effort and uncertainty serve as robust predictors of both final and dynamic privacy risk.
}
\end{kkboxline}

\begin{table*}[t]
    \centering
    \tiny
    \setlength{\tabcolsep}{1.0pt}
    \caption{\textbf{Correlation between sample hardness metrics and membership vulnerability.} Pearson correlation coefficients are shown for both static vulnerability ($\alpha$) and dynamic vulnerability ($v_{\alpha}$). \textbf{Bold} values denote $|r| \ge 0.80$ (very strong correlation).}
    \label{tab:merged_corr}
    \scalebox{1.4}{%
      \begin{tabular}{@{}p{1.8cm}>{\columncolor{gray!15}\raggedright\arraybackslash}p{1.9cm} l ccc>{\columncolor{lightgray}}c>{\columncolor{lightgray}}c>{\columncolor{lightgray}}c ccc@{}}
        \toprule
        \textbf{Group} & \textbf{Metric} & \textbf{Target} & \multicolumn{3}{c}{\textbf{CNN-32}} & \multicolumn{3}{c}{\cellcolor{lightgray}\textbf{CNN-64}} & \multicolumn{3}{c}{\textbf{WRN-28}} \\
        \cmidrule(lr){4-6} \cmidrule(lr){7-9} \cmidrule(lr){10-12}
        \multicolumn{3}{c}{} & All & $\alpha>0$ & $\alpha\le0$ & All & $\alpha>0$ & $\alpha\le0$ & All & $\alpha>0$ & $\alpha\le0$ \\
        \midrule
        \multirow{2}{*}{\textit{Low-level Opt.}} & \multirow{1}{*}{Gradient Norm} & $\alpha$ & 0.000 & 0.005 & $-0.012$ & 0.003 & 0.004 & $-0.012$ & 0.002 & 0.001 & 0.001 \\
        &  & $v_{\alpha}$ & 0.000 & 0.007 & $-0.005$ & 0.003 & 0.005 & $-0.008$ & 0.004 & 0.008 & 0.002 \\
        \midrule
        \multirow{4}{*}{\textit{Cumulative Effort}} & Iteration Learned & $\alpha$ & 0.411 & 0.708 & $-0.613$ & 0.559 & 0.755 & $-0.624$ & 0.564 & 0.692 & $-0.493$ \\
        &  & $v_{\alpha}$ & 0.402 & 0.674 & $-0.593$ & 0.548 & 0.734 & $-0.615$ & 0.565 & 0.719 & $-0.554$ \\[0.15em]
        & Influence Function & $\alpha$ & 0.450 & 0.680 & $-0.600$ & 0.510 & 0.740 & $-0.630$ & 0.480 & 0.710 & $-0.460$ \\
        &  & $v_{\alpha}$ & 0.425 & 0.690 & $-0.580$ & 0.500 & 0.720 & $-0.640$ & 0.520 & 0.740 & $-0.520$ \\
        \midrule
        \multirow{4}{*}{\textit{Uncertainty}} & Aleatoric & $\alpha$ & 0.500 & 0.778 & $-0.659$ & 0.654 & \textbf{0.812} & $-0.647$ & 0.779 & \textbf{0.885} & $-0.642$ \\
        &  & $v_{\alpha}$ & 0.488 & 0.742 & $-0.619$ & 0.635 & 0.786 & $-0.633$ & 0.776 & \textbf{0.913} & $-0.764$ \\[0.15em]
        & Epistemic & $\alpha$ & 0.505 & 0.766 & $-0.633$ & 0.656 & \textbf{0.801} & $-0.612$ & \textbf{0.808} & \textbf{0.908} & $-0.609$ \\
        &  & $v_{\alpha}$ & 0.494 & 0.737 & $-0.591$ & 0.636 & 0.777 & $-0.597$ & 0.790 & \textbf{0.916} & $-0.725$ \\
        \bottomrule
      \end{tabular}%
    }
    \vspace{-10pt}
\end{table*}

\noindent
\textbf{Description.}
To understand why certain individual samples are more vulnerable than others, we examine the relationship between a sample’s intrinsic “hardness”—the difficulty a model encounters when learning it—and susceptibility to membership inference. We quantify hardness using three categories of metrics. First, low-level optimization signals, such as Gradient Norm~\cite{chen2018gradnormgradientnormalizationadaptive}, indicate the magnitude of parameter updates per sample. Second, cumulative learning effort metrics include Iteration Learned~\cite{toneva2019empiricalstudyexampleforgetting}, which tracks how long a sample takes to be consistently learned, and Influence Functions~\cite{koh2020understandingblackboxpredictionsinfluence}, measuring a sample’s impact on model predictions. Third, model uncertainty metrics, specifically Aleatoric and Epistemic Uncertainty~\cite{hullermeier2021aleatoric}, assess the model’s confidence or lack thereof, with high epistemic uncertainty typically linked to samples challenging to generalize and thus at greater risk of memorization.

We compute the Pearson correlation between these hardness metrics and two dimensions of privacy risk: the static membership advantage ($\alpha$) after training, and the dynamic membership encoding speed ($v_\alpha$) that quantifies how rapidly privacy vulnerability evolves.

\noindent
\textbf{Results.}
Table~\ref{tab:merged_corr} summarizes correlation findings, revealing distinct predictive power across different hardness metrics. Low-level optimization signals like gradient norm exhibit negligible correlations with both static and dynamic vulnerability, indicating privacy leakage arises from cumulative learning trajectories rather than isolated parameter updates.

In contrast, cumulative learning effort metrics—iteration learned and influence functions—show strong positive correlations with privacy risk. Model uncertainty metrics, especially epistemic uncertainty, demonstrate an exceptionally high correlation with vulnerability (0.916 for dynamic vulnerability in the WRN-28 model). This indicates that high epistemic uncertainty, reflecting the model's struggle to generalize, significantly increases memorization risk.

Conditioning on the final vulnerability status offers additional insights. For the vulnerable subset ($\alpha > 0$), the positive correlation between learning difficulty and privacy risk is particularly pronounced, emphasizing that harder-to-learn samples incur greater leakage. Conversely, within the non-vulnerable subset ($\alpha \leq 0$), negative correlations suggest that challenging samples in this group enhance model generalization rather than memorization, reducing privacy risks. Despite this nuanced subset behavior, the overall correlation remains positive, primarily driven by the strong relationship observed in the vulnerable subgroup.

\subsection{Early Exposure of Vulnerable Samples}

\begin{kkboxline}
\textbf{Finding 5:}
\textit{
Vulnerability trajectories are not random; they are governed by sample hardness, and their ultimate direction is often established early in training, creating a critical window for proactive intervention.
}
\end{kkboxline}


\noindent
\textbf{Description.}
To investigate when privacy risk emerges during training, we trace the vulnerability trajectory, $\tau(z)$, for each sample on the vulnerability plane. To quantify the magnitude of a sample's vulnerability evolution, we define a Vulnerability Path Length, which measures the cumulative change in a sample's membership advantage score over time. For a sample $z$, its advantage at epoch $t$ is $\alpha_t(z) = \mathrm{TPR}_t(z) - \mathrm{FPR}_t(z)$. The total vulnerability path length, $L(z)$, is the sum of the absolute changes in advantage across all training intervals:

\begin{equation}
    L(z) = \sum_{t=0}^{T-1} |\alpha_{t+1}(z) - \alpha_t(z)|
\end{equation}

This metric rewards movement perpendicular to the non-vulnerable diagonal (i.e., changes in risk) while penalizing movement parallel to it (where risk remains constant).
Using this measure, we stratify the dataset by identifying ``high-travel" samples, which exhibit the longest total path length on the plane and thus the most substantial changes in vulnerability, and ``low-travel" samples, which have the shortest total path length. We then analyze the distinct learning characteristics, such as loss distributions and hardness scores, of these two groups at early, middle, and late stages of training. 

\begin{figure}[h]
        \centering
        \includegraphics[width=0.9\columnwidth]{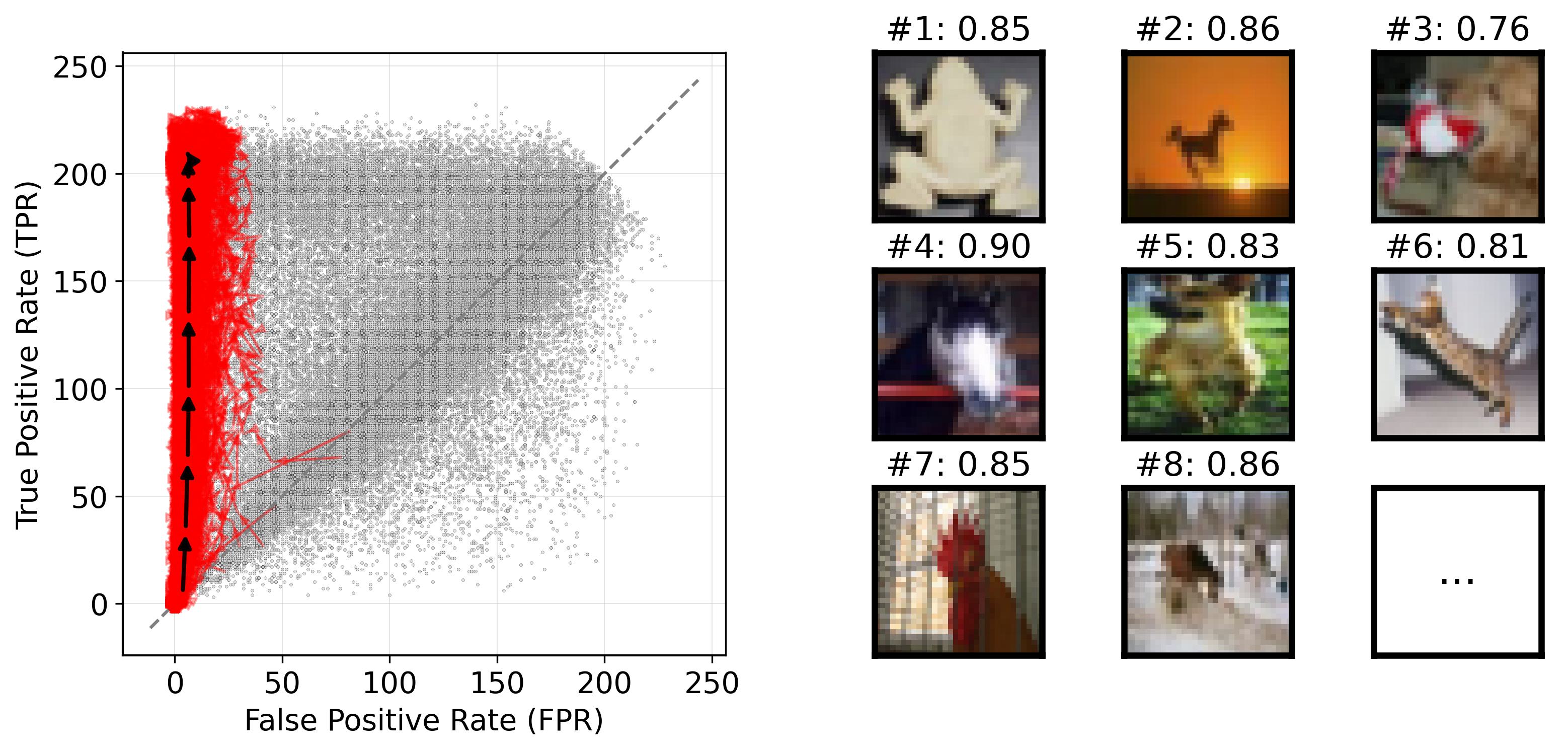}
        \caption{Correlation between sample-level membership advantage trajectory and sample epistemic uncertainty score. Red arrows indicate the trajectory of the top 100 most vulnerable samples at 10-epoch intervals, while the black arrow shows the average trajectory for all vulnerable samples over 40 epochs.} 
        \label{fig:score_traj_comparison}
\end{figure}

\noindent
\textbf{Results.}
Our analysis reveals that the dynamics of membership encoding are not arbitrary but are strongly dictated by the intrinsic properties of each sample. Figure~\ref{fig:score_traj_comparison} illustrates a direct correlation where samples with the longest and most volatile vulnerability trajectories—the ``high-travel" samples—are consistently those with the highest learning difficulty, as quantified by epistemic uncertainty. In contrast, ``low-travel" samples are typical examples learned easily and stably. This confirms that the geometric trajectory a sample follows through the vulnerability state space directly reflects the model’s difficulty in learning that sample.

\begin{figure}[t!]
    \centering
    \begin{subfigure}[t]{0.48\columnwidth}
        \centering
        \includegraphics[width=0.9\columnwidth]{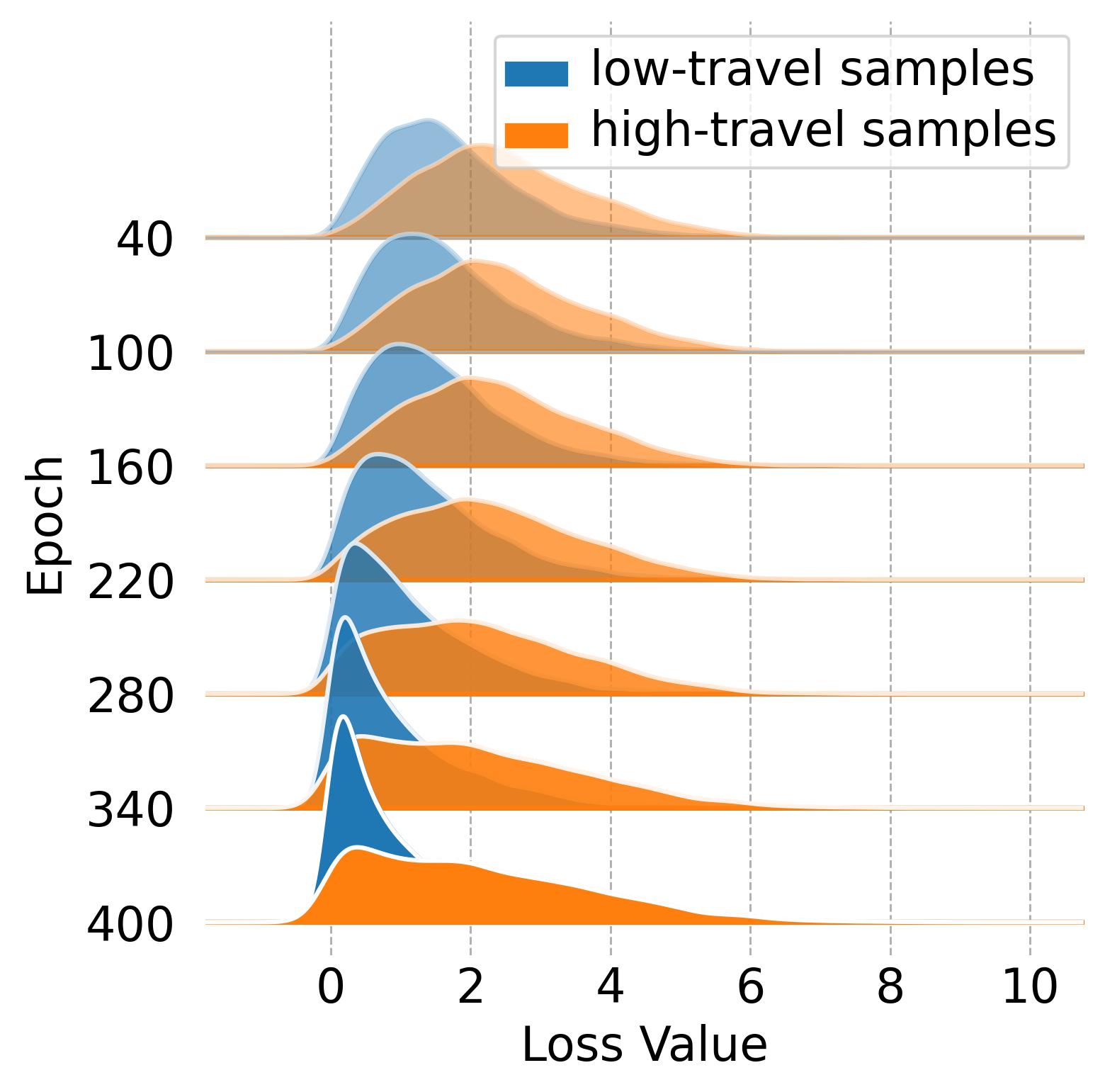}
        \caption{Loss distribution changes for shortest travel points and longest travel points at different training epochs.} 
        \label{fig:loss_distribution}
    \end{subfigure}
    \hfill
    \begin{subfigure}[t]{0.48\columnwidth}
        \centering
        \includegraphics[width=0.9\columnwidth]{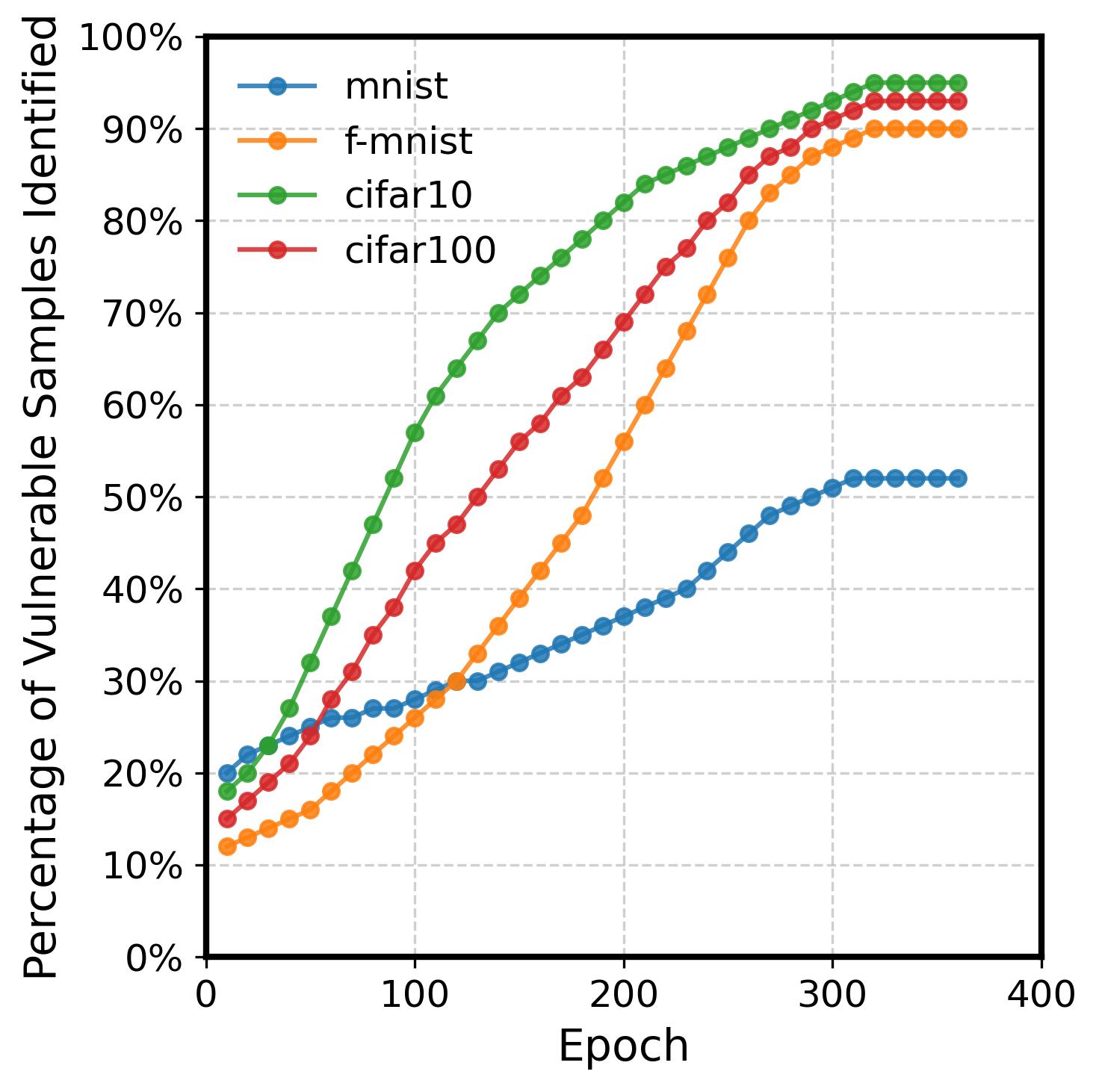}
        \caption{Percentage of vulnerable samples identified over epoch under different datasets}
        \label{fig:vulnerable_sample_dataset_epoch_analyze}
    \end{subfigure}%

    \caption{Early predictability of membership vulnerability.}
    \label{fig:vul_datasets_analyze_score}
\end{figure}

This fundamental distinction between hard (high-travel) and easy (low-travel) samples emerges notably early in the training process. As shown in Figure~\ref{fig:loss_distribution}, while the loss distributions of these two groups initially overlap, they diverge significantly as training progresses. Specifically, the loss for high-travel samples remains consistently high, indicating ongoing model difficulty, whereas the loss for low-travel samples quickly converges. Their early divergence during training raises a critical question: how early is a sample's vulnerability to membership inference actually determined?


The answer is, remarkably early. At any given stage of training, we can flag the samples with the longest vulnerability path lengths up to that point as high-risk. Figure~\ref{fig:vulnerable_sample_dataset_epoch_analyze}, which plots the cumulative portion of the final vulnerable population that is exposed over time, offers a clear picture. The curve for complex datasets like CIFAR-10 rises sharply and then begins to plateau, revealing a pattern of rapid, front-loaded encoding. This occurs because the model, in its early, aggressive learning phase, quickly resorts to memorization for difficult examples it cannot easily generalize, causing their membership information to be exposed. Our analysis shows this exposure happens swiftly; over 70\% of all samples that will ultimately be vulnerable in the final model have already revealed this predisposition by just epoch 150.

\section{Discussion}
This work reframes membership privacy analysis, moving from a static, post-hoc audit to a dynamic, in-training process. We demonstrate that privacy leakage is not a random failure but a predictable consequence of a model’s struggle to generalize. This struggle is most acute for specific ``hard-to-learn" samples, and their vulnerability trajectory is strongly correlated with measurable learning metrics like epistemic uncertainty. By tracking these dynamics on a vulnerability plane, we can visualize how factors like dataset complexity and model architecture create a ``privacy-underprivileged" subset of data that the model is forced to memorize.

The central implication of this work is the shift from reactive defense to proactive privacy engineering. Our discovery that a sample's vulnerability is effectively sealed within a predictable early window fundamentally changes the strategic landscape. This opens the door to move beyond costly, uniform protections and toward designing targeted, efficient interventions that act upon the small fraction of samples that exhibit signatures of high learning difficulty during this critical phase. The ability of optimizers like Sharpness-Aware Minimization (SAM) to inherently suppress these vulnerability trajectories proves the principle. Ultimately, by understanding how membership is encoded, we can design learning algorithms that are not just audited for privacy, but are actively guided toward it during their creation, making them private by design.
\section{Conclusion}
This paper introduced a novel framework to dissect the dynamics of membership privacy at the individual sample level, moving beyond static, post-hoc analysis. Our analysis reveals that membership information encoding is a predictable process governed by the interplay of dataset complexity, model architecture, and optimizer choice. We demonstrated that a sample’s vulnerability is strongly correlated with its intrinsic learning difficulty, which can be quantified by metrics like epistemic uncertainty during training. Crucially, these high-risk samples are exposed early in the training process, within a critical window where intervention is most effective. 

\bibliography{aaai2026}

\appendix
\section{Notation Table}

Table~\ref{tab:notation} provides a summary of the mathematical notations used throughout the paper for quick reference.

\section{Consistency of Findings with Alternative MIA Methods}
\label{app:consistency}
\begin{table}[h]
    \centering
    \small
    \setlength{\tabcolsep}{3pt}
    \caption{Performance metrics across datasets using the \citet{shokri2017membership} attack (subscripts indicate standard deviation; values with $^{\dagger}$ scaled by $10^{-3}$).}
    \label{tab:appendix_metrics_data}
    \resizebox{\linewidth}{!}{%
      \begin{tabular}{@{}l>{\columncolor{gray!15}}lcccc@{}}
        \toprule
        \textbf{Group} & \textbf{Metric} & \textbf{MNIST} & \textbf{F-MNIST} & \textbf{CIFAR10} & \textbf{CINIC10} \\
        \midrule
        \multirow{3}{*}{\textit{Motion}} & Speed ($\bar{v}$)$^{\dagger}$   & 0.25\textsubscript{0.30} & 0.95\textsubscript{0.61} & 2.15\textsubscript{1.10} & 2.80\textsubscript{1.15} \\
                                         & $\alpha{>}0$ Speed$^{\dagger}$ & 0.45\textsubscript{0.60} & 0.70\textsubscript{0.52} & 1.80\textsubscript{0.95} & 4.10\textsubscript{1.05} \\
                                         & CoM Disp.                     & 9.5                     & 35.2                    & 85.3                     & 102.2                    \\
        \midrule
        \multirow{3}{*}{\textit{Information}} & Entropy ($H$)    & 4.38\textsubscript{0.06} & 4.25\textsubscript{0.40} & 4.61\textsubscript{0.72} & 4.75\textsubscript{0.55} \\
                                              & $\Delta H$       & \(-0.05\)               & 1.10                    & 2.95                     & 3.68                     \\
                                              & Dir. Angle ($\theta$) & \(-0.65\)           & 0.98                    & 0.88                     & 0.92                     \\
        \midrule
        \multirow{2}{*}{\textit{Clustering}} & Avg.\ ($\bar{k}$) & 3250\textsubscript{30}  & 2410\textsubscript{350} & 1890\textsubscript{400} & 1650\textsubscript{380} \\
                                            & $\Delta k$        & 4                       & 1205                    & 1050                    & 1120                     \\
        \bottomrule
      \end{tabular}%
    }
\end{table}

To validate that our dynamic framework captures fundamental properties of the learning process rather than artifacts of a specific Membership Inference Attack (MIA), we systematically replicated our main experiments using the foundational shadow-modeling attack by \citet{shokri2017membership}. This attack, which employs binary classifiers trained on shadow model outputs to distinguish members from non-members, serves as a methodologically distinct alternative to the likelihood-ratio-based LiRA~\cite{lira}. While the \citet{shokri2017membership} attack is a generally weaker adversary, yielding lower absolute vulnerability scores, our analysis reveals that it uncovers identical qualitative trends and relative dynamics, thus confirming the robustness of our conclusions.


\begin{kkboxline}
\textbf{Consistent Finding:}
\textit{
The core dynamic trends—such as the influence of dataset complexity and model architecture on vulnerability trajectories—are consistent regardless of the underlying MIA method used for measurement.
}
\end{kkboxline}

\mypara{Impact of Dataset Complexity}
We first re-evaluated the impact of dataset complexity by training a `wrn28-2` model on MNIST, F-MNIST, CIFAR-10, and CINIC-10, measuring the dynamics using the~\citet{shokri2017membership} attack. Table~\ref{tab:appendix_metrics_data} presents the results.

By comparing these results with those from LiRA (Table~\ref{tab:evaluation_metrics_data}), we observe identical trends. Although the absolute values for metrics like CoM Displacement and Change in Spatial Entropy ($\Delta H$) differ, they both increase monotonically with dataset complexity. For instance, the CoM Displacement measured by the Shokri attack increases from 9.5 (MNIST) to 102.2 (CINIC-10). This confirms our primary finding that more complex datasets induce a significantly greater and more heterogeneous encoding of membership information, a phenomenon detectable by fundamentally different attack methodologies.

\begin{table}[h]
    \centering
    \small
    \setlength{\tabcolsep}{3pt}
    \caption{Metrics on different models using the \citet{shokri2017membership} attack (subscripts indicate standard deviation; values with $^{\dagger}$ scaled by $10^{-3}$).}
    \label{tab:appendix_metrics_arch}
    \resizebox{\linewidth}{!}{%
      \begin{tabular}{@{}l>{\columncolor{gray!15}}lccc@{}}
        \toprule
        \textbf{Group} & \textbf{Metric} & \textbf{cnn32-3-max} & \textbf{cnn64-3-max} & \textbf{wrn28-2} \\
        \midrule
        \multirow{3}{*}{\textit{Motion}} & Speed ($\bar{v}$)$^{\dagger}$   & 0.80\textsubscript{0.55} & 2.05\textsubscript{1.00} & 2.15\textsubscript{1.10} \\
                                         & $\alpha{>}0$ Speed$^{\dagger}$ & 1.10\textsubscript{0.65} & 1.75\textsubscript{0.90} & 1.80\textsubscript{0.95} \\
                                         & CoM Disp.                     & 28.9                    & 80.1                     & 85.3                     \\
        \midrule
        \multirow{3}{*}{\textit{Information}} & Entropy ($H$)    & 4.35\textsubscript{0.15} & 4.58\textsubscript{0.65} & 4.61\textsubscript{0.72} \\
                                              & $\Delta H$       & 0.25                    & 2.89                    & 2.95                     \\
                                              & Dir. Angle ($\theta$) & \(-0.42\)           & 0.85                    & 0.88                     \\
        \midrule
        \multirow{2}{*}{\textit{Clustering}} & Avg.\ ($\bar{k}$) & 2250\textsubscript{150.5} & 1815\textsubscript{380.1} & 1890\textsubscript{400} \\
                                            & $\Delta k$        & 650                      & 980                      & 1050                     \\
        \bottomrule
      \end{tabular}%
    }
\end{table}

\begin{table*}[t]
    \centering
    \begin{tabular}{l l}
    \toprule
    \textbf{Symbol} & \textbf{Description} \\
    \midrule
    \multicolumn{2}{l}{\textbf{General Machine Learning}} \\
    $\mathcal{X}, \mathcal{Y}$ & Input feature space and label space, respectively. \\
    $\mathcal{Z}$ & The complete data space, $\mathcal{Z} = \mathcal{X} \times \mathcal{Y}$. \\
    $z, z^*$ & A data sample $(x, y)$, with $z^*$ denoting a specific target sample. \\
    $\mathcal{D}$ & The underlying data distribution. \\
    $D_{train}, D_{out}, D_{aux}$ & The training set, holdout (non-member) set, and auxiliary set, respectively. \\
    $n$ & The number of samples in the training set. \\
    $\mathcal{M}_\theta$ & A machine learning model with parameters $\theta$. \\
    $\Theta$ & The parameter space for the model. \\
    $\theta_t$ & The model's parameters at training epoch $t$. \\
    $T$ & The total number of training epochs. \\
    $f_{\mathcal{M}_\theta}(x')$ & The output vector (e.g., logits, probabilities) of model $\mathcal{M}_\theta$ for input $x'$. \\
    $\mathcal{L}$ & The loss function. \\
    $\nabla_{\theta_t} \mathcal{L}$ & The gradient of the loss function with respect to parameters $\theta_t$. \\
    \midrule
    \multicolumn{2}{l}{\textbf{Membership Inference Attacks (MIAs)}} \\
    $\mathcal{A}$ & The adversary in the context of the MIA game. \\
    $\mathcal{T}$ & The training algorithm that maps a dataset to model parameters. \\
    $\mathcal{F}$ & A membership inference attack function or statistical test. \\
    $H_1, H_0$ & The \textit{in}-hypothesis ($z \in D_{train}$) and \textit{out}-hypothesis ($z \notin D_{train}$). \\
    $\mathcal{P}_{\mathrm{in}}, \mathcal{P}_{\mathrm{out}}$ & The distribution over model parameters when $z$ is included or excluded. \\
    $\mathrm{TPR}_{z}, \mathrm{FPR}_{z}$ & The True Positive Rate and False Positive Rate of an MIA for sample $z$. \\
    $\mathrm{Adv}^{\mathcal{F}}(z)$ & The theoretical membership advantage of attack $\mathcal{F}$ on sample $z$. \\
    $\alpha_{z}$ & The empirical membership advantage for sample $z$, calculated as $\mathrm{TPR}_{z} - \mathrm{FPR}_{z}$. \\
    \midrule
    \multicolumn{2}{l}{\textbf{Dynamic Analysis Framework}} \\
    $\mathcal{V}$ & The 2D vulnerability plane, defined by the unit square $[0, 1] \times [0, 1]$. \\
    $v_t(z)$ & The vulnerability state of sample $z$ at epoch $t$, given by $(\mathrm{FPR}_{z}, \mathrm{TPR}_{z})$. \\
    $\tau(z)$ & The vulnerability trajectory of sample $z$ over training, $(v_1(z), \dots, v_T(z))$. \\
    $\vec{s}(z, t)$ & The membership encoding velocity (discrete derivative) of $z$'s state at epoch $t$. \\
    $\mathcal{S}$ & A finite set of discretized states on the vulnerability plane $\mathcal{V}$. \\
    $A(t)$ & The transition matrix between vulnerability states at time step $t$. \\
    $\mu_t$ & The empirical distribution of the entire sample population on the plane at epoch $t$. \\
    $\mathrm{CoM}_t$ & The Center of Mass of the sample population on the vulnerability plane at epoch $t$. \\
    $H(\mu_t)$ & The spatial entropy of the sample distribution on the vulnerability plane at epoch $t$. \\
    $v_\alpha$ & The membership encoding speed, capturing the rate of change in advantage $\alpha$. \\
    $L(z)$ & The Vulnerability Path Length of sample $z$, $\sum |\alpha_{t+1}(z) - \alpha_t(z)|$. \\
    \bottomrule
    \end{tabular}
    \caption{Notation Table}
    \label{tab:notation}
\end{table*}

\mypara{Impact of Dataset Complexity}
Our investigation using the \citet{shokri2017membership} attack corroborates the finding that dataset complexity is a primary driver of privacy leakage dynamics. As detailed in Table~\ref{tab:appendix_metrics_data}, the dynamic metrics measured with this weaker attack show the same monotonic increase with dataset complexity as observed with LiRA (Table~\ref{tab:evaluation_metrics_data}). For instance, both the total path length of the population's center of mass (\textbf{CoM Displacement}) and the increase in risk heterogeneity (\textbf{Change in Spatial Entropy, $\boldsymbol{\Delta H}$}) consistently grow when moving from MNIST to CINIC-10. This consistency demonstrates that the underlying geometric shift in the vulnerability space is an invariant property of the learning task, detectable even by less powerful adversaries.

\mypara{Impact of Model Architecture}
The framework's conclusion that model capacity acts as a catalyst for membership encoding is also robust to the choice of attack. The results in Table~\ref{tab:appendix_metrics_arch}, obtained using the \citet{shokri2017membership} attack on CIFAR-10, align perfectly with the LiRA-based findings in Table~\ref{tab:evaluation_metrics_arch}. Increasing model capacity from a shallow `cnn32-3-max' to a deep `wrn28-2' results in a clear and consistent increase across all \textit{Motion}, \textit{Information}, and \textit{Clustering} metrics. This confirms that higher-capacity models inherently induce a more pronounced bifurcation of samples into secure and vulnerable populations, a structural dynamic that our framework reliably captures independent of the specific MIA used as a measurement tool.

\begin{figure}[h]
    \centering
    \includegraphics[width=\columnwidth]{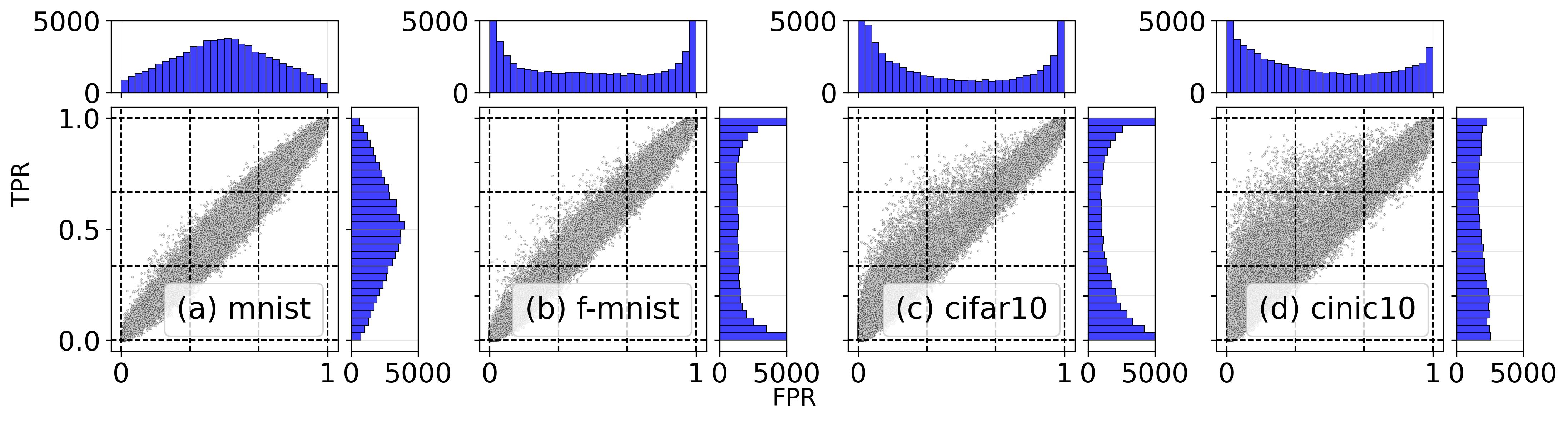}
    \caption{Post-training FPR-TPR planes for different datasets: (a) MNIST, (b) Fashion MNIST, (c) CIFAR-10, and (d) CINIC-10, measured by the ~\citet{shokri2017membership} attack.}
    \label{fig:appendix_plane_comparison_dataset}
\end{figure}

\begin{figure}[h]
    \centering
    \includegraphics[width=0.95\columnwidth]{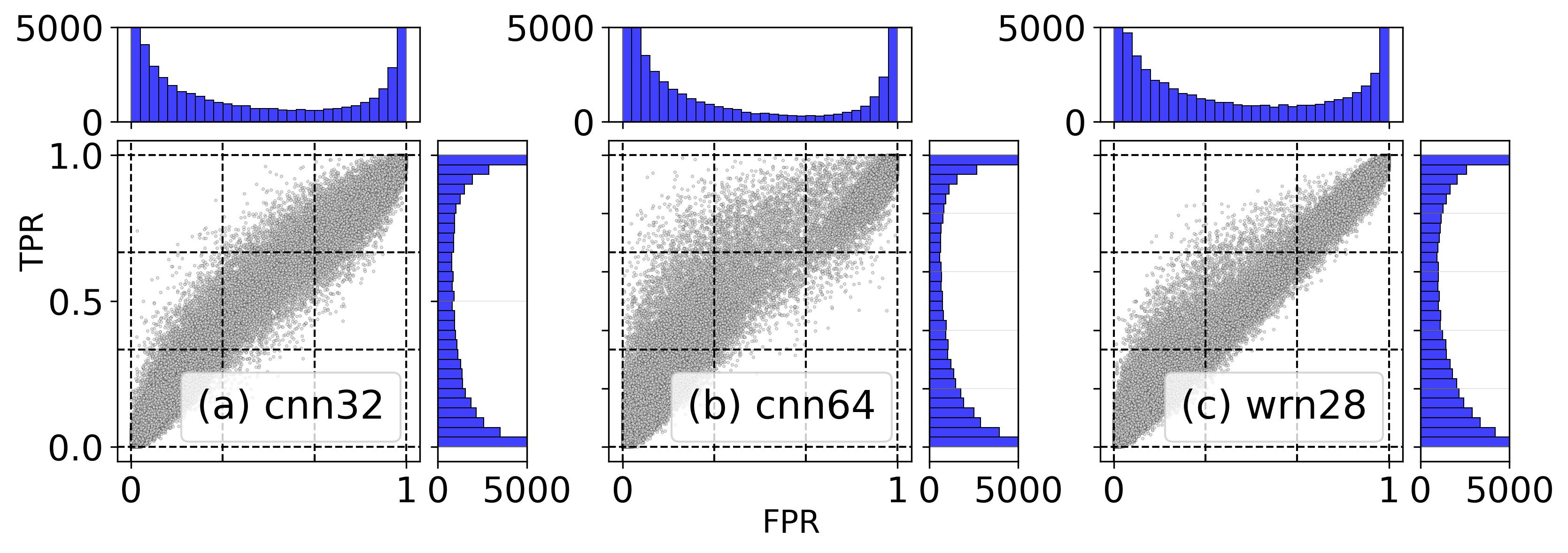}
    \caption{Post-training FPR-TPR planes for different architectures: (a) cnn32-3-max, (b) cnn64-3-max, and (c) wrn28-2, measured by the ~\citet{shokri2017membership} attack.}
    \label{fig:appendix_plane_comparison_arch}
\end{figure}

\begin{figure}[h]
    \centering
    \includegraphics[width=0.7\columnwidth]{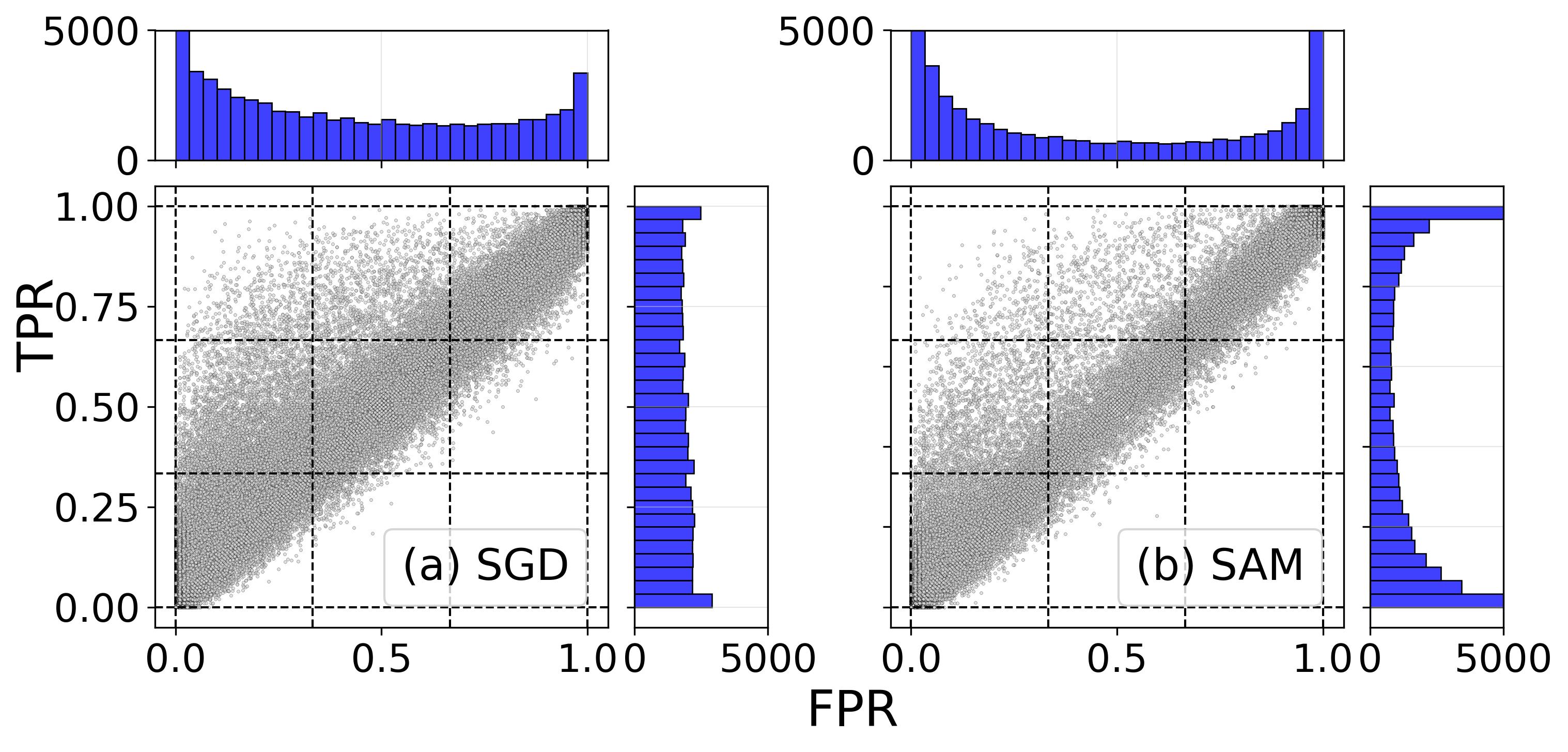}
    \caption{Post-training FPR-TPR planes for different optimizers: (a) SGD, (b) SAM, measured by the ~\citet{shokri2017membership} attack.}
    \label{fig:appendix_plane_comparison_optim}
\end{figure}

\mypara{Visual Confirmation of Dynamic Consistency}
The visual evidence supporting our findings also remains consistent across different MIAs. We reproduced the key visualizations from the main text using the vulnerability scores derived from the \citet{shokri2017membership} attack.

First, the post-training distributions on the Vulnerability Plane exhibit the same structural changes as we seen in Figure~\ref{fig:datasets_analyze}. As shown in Figure~\ref{fig:appendix_plane_comparison_dataset},~\ref{fig:appendix_plane_comparison_arch}, and~\ref{fig:appendix_plane_comparison_optim}, while the ~\citet{shokri2017membership} attack yields generally lower TPR and higher FPR values (shifting the entire distribution down and to the right), the core trend is preserved. For simpler datasets like MNIST, the sample population remains tightly clustered along the non-vulnerable diagonal. For complex datasets like CINIC-10, the distribution disperses significantly, populating the upper regions of the plane and confirming a more heterogeneous risk landscape.

Second, the Transition Probabilities from the robust region ($S_{11}$) to the highly vulnerable region ($S_{31}$) follow identical temporal patterns, analogous to Figure~\ref{fig:trans_mat}. Although the peak probability measured by the ~\citet{shokri2017membership} attack is lower due to its weaker signal, the relative ordering and the timing of the ``tipping point" are consistent. For example, the transition probability for CINIC-10 still peaks earlier and at a higher relative magnitude than for CIFAR-10. This demonstrates that the existence of a critical window for membership encoding is a fundamental property of the training dynamics, not an artifact of LiRA's sensitivity.

\end{document}